\PassOptionsToPackage{table}{xcolor}
\documentclass[10pt,twocolumn,letterpaper,pagenumbers]{article}

\usepackage{iccv}              

\definecolor{iccvblue}{rgb}{0.21,0.49,0.74}
\usepackage[pagebackref,breaklinks,colorlinks,allcolors=iccvblue]{hyperref}

\definecolor{Gray}{gray}{0.92}
\definecolor{codegreen}{rgb}{0.0,0.6,0.0}
\usepackage[ruled,vlined,linesnumbered,noend]{algorithm2e}
\makeatletter
\newcommand{\algorithmfootnote}[2][\footnotesize]{%
  \let\old@algocf@finish\@algocf@finish
  \def\@algocf@finish{\old@algocf@finish
    \leavevmode\rlap{\begin{minipage}{\linewidth}
    #1#2
    \end{minipage}}%
  }%
}
\usepackage{diagbox}
\usepackage{multirow}
\renewcommand{\footnotesize}{\scriptsize}
\makeatother

\def\paperID{2037} 
\def\confName{ICCV}
\def\confYear{2025}

\title{From Slices to Sequences: Autoregressive Tracking Transformer for Cohesive and Consistent 3D Lymph Node Detection in CT Scans}

\author{Qinji Yu$^{1,2\thanks{Equal contribution.}}$\,\, Yirui Wang$^{1\textsuperscript{\thefootnote}}$ \,\, Ke Yan$^{1,4}$ \,\, Dandan Zheng$^{3}$ \,\, Dashan Ai$^{5}$ \,\, Dazhou Guo$^{1}$ \\ Zhanghexuan Ji$^{1}$ \,\, Yanzhou Su$^{1,4}$\,\, Yun Bian$^{6}$ \,\,  Na Shen$^{7}$ \,\, Xiaowei Ding$^{2}$ \,\,  Kuaile Zhao$^{5}$ \,\, Le Lu$^{1}$ \\ Xianghua Ye$^{3}$ \,\, Dakai Jin$^{1}$\\ 
\vspace{-3mm}
\\  
$^1$DAMO Academy, Alibaba Group \,\, $^2$ Shanghai Jiao Tong University \\ $^3$The First Affiliated Hospital of Zhejiang University \,\, $^4$Hupan Lab, 310023, Hangzhou, China \\ $^5$Fudan University Shanghai Cancer Center \,\, $^6$Changhai Hospital \,\, $^7$Zhongshan Hospital, Fudan University
}

\begin{document}
\maketitle
\begin{abstract}
Lymph node (LN) assessment is an essential task in the routine radiology workflow, providing valuable insights for cancer staging, treatment planning and beyond. Identifying scatteredly-distributed and low-contrast LNs in 3D CT scans is highly challenging, even for experienced clinicians. Previous lesion and LN detection methods demonstrate effectiveness of 2.5D approaches (i.e, using 2D network with multi-slice inputs), leveraging pretrained 2D model weights and showing improved accuracy as compared to separate 2D or 3D detectors. However, slice-based 2.5D detectors do not explicitly model inter-slice consistency for LN as a 3D object, requiring heuristic post-merging steps to generate final 3D LN instances, which can involve tuning a set of parameters for each dataset. In this work, we formulate 3D LN detection as a tracking task and propose LN-Tracker, a novel LN tracking transformer, for joint end-to-end detection and 3D instance association. Built upon DETR-based detector, LN-Tracker decouples transformer decoder's query into the track and detection groups, where the track query autoregressively follows previously tracked LN instances along the z-axis of a CT scan. We design a new transformer decoder with masked attention module to align track query's content to the context of current slice, meanwhile preserving detection query's high accuracy in current slice. An inter-slice similarity loss is introduced to encourage cohesive LN association between slices. Extensive evaluation on four lymph node datasets shows LN-Tracker's superior performance, with at least $2.7\%$ gain in average sensitivity when compared to other top 3D/2.5D detectors. Further validation on public lung nodule and prostate tumor detection tasks confirms the generalizability of LN-Tracker as it achieves top performance on both tasks.
\end{abstract}    
\vspace{-0.4em}
\section{Introduction}
\label{sec:intro}

\begin{figure}[t]
\centering
\includegraphics[width=1.0\columnwidth]{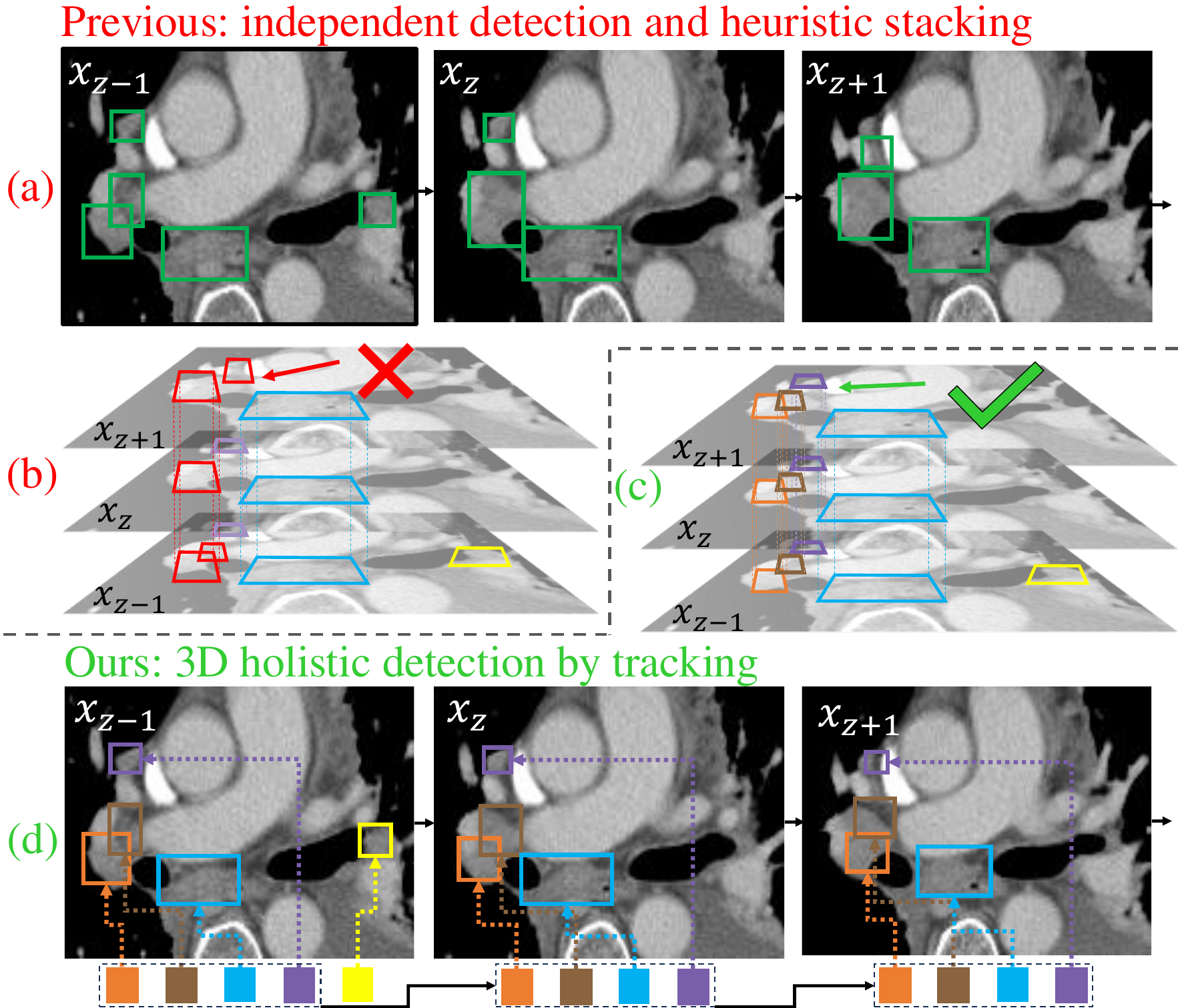}
\caption{3D detection results by 2.5D detector with heuristic stacking methods (a,b)~\cite{yan2020learning,yu2025effective}, and by the proposed 3D detection-by-tracking (c,d). Due to the lack of explicitly modeling of inter-slice consistency and heuristic post-merging of 2.5D methods, 3D LNs are falsely divided to two individual instances or merged to a larger instance as indicated by \textcolor{red}{red boxes} in (b). }
\label{Fig:track_motivation}
\vspace{-1.5em}
\end{figure}

As essential components of the human immune system, lymph nodes (LNs) are widely distributed throughout the body along lymphatic pathways and serve as primary routes for cancer spread. Therefore, LN detection and measurement is one of the major tasks in the daily radiology and oncology workflow, which is essential for cancer staging, treatment planning, and prognostic analysis~\cite{takeuchi2009validation,chang2020axillary,jin2022towards}. CT is the primary imaging modality for preoperative LN diagnosis. However, due to the imaging limitations, such as low contrast and coarse spatial resolution, LNs often appear with indistinct intensity and similar size/shape to adjacent soft tissues. Consequently, even experienced clinicians struggle to identify all clinically relevant LNs in a volumetric CT scan, which usually comprises hundreds of 2D slices, within a limited time. Studies have shown low accuracy in mediastinal LN diagnosis ($40\%\sim65\%$)~\cite{mcloud1992bronchogenic,backhus2013radiographic}, showing the demand for accurate Computer-aided detection (CADe) solutions.



3D detector is a natural choice for lesion detection in medical imaging, since 3D deep network architecture can simultaneously capture spatial context across all three dimensions. 
While straightforward, 3D detectors for LN detection can suffer from two major issues: (1) diagnostic CT scans can be \textit{extremely anisotropic in 3D, \ie, z-spacing$\gg$x,y-spacing}. As shown in Table \ref{tab:dataset}, CT may have 4-5mm z-spacing, which is {6 to 10 times} greater than the x,y-spacing of [0.46, 0.70]mm. In contrast, \textit{size of LN is small}, usually ranging from 3mm to 20mm, leading to large variations in LN appearance between adjacent z-slices and meanwhile many LNs only possess 2-3 slices. This makes 3D methods difficult/ineffective to integrate or fuse spatial LN features. (2) 3D detectors lack pre-trained 3D model weights, which requires more training data to be fully optimized.

To conquer the limitation of the 3D detector, recent works mainly explore the 2.5D approach~\cite{yan2019mulan,yan2020learning,yang2021asymmetric,wang2022global,zhao2023diffuld,yu2025effective}. 2.5D configuration first extracts feature maps of the input sequence independently via a pre-trained 2D network. Then, the target-slice features are enhanced by cross-slice feature fusion~\cite{yan2019mulan,li2022satr,zhang2023advancing} and used to predict the target-slice result. After that, a heuristic merging process is required to combine 2D individual predictions into 3D instances~\cite{yan2020learning,yan2023anatomy,yu2025effective,yu2024slice}. Improved detection results are reported than 3D detectors. However, slice-based 2.5D detectors do not explicitly model lesion's inter-slice consistency, and require carefully (often ad hocly) tuning of parameters in the post-merging step according to the specific task and dataset, which lead to under- or over-merged 3D instances significantly limiting its generalizability in clinical usage. For LN detection, it is often observed that a single 3D LN can be falsely predicted as two or more instances or multiple LNs are erroneously merged into one large LN~\cite{yan2020learning,yu2025effective} (Figure~\ref{Fig:track_motivation}). These will adversely affect the downstream LN metastasis diagnosis because 3D size information is one of the most important LN malignant indicators.

The inherent limitation of 2.5D detector motivates us to tackle 3D LN detection from a new perspective. Inspired by the diagnosis process of radiologists, who identify  LNs in a CT scan by frequently examining adjacent slices to evaluate the continuity and similarity, we \textit{formulate 3D LN detection as a multi-object tracking task, i.e., slice-by-slice detection with cross-slice instance association}. Under the detection-by-tracking formulation, detection backbone can be a 2.5D architecture to inherit the advantage of 2.5D-based detector. Meanwhile, the tracking formulation explicitly enforces spatial-temporal constraints to allow 3D LN instance information from the previous slice to propagate to the current slice, promoting detection cohesion and consistency without the post-merging heuristic process~\cite{yan2020learning,yan2023anatomy,yu2025effective}, which solves the major issues of 2.5D detector.

\begin{figure}[t]
\centering
\includegraphics[width=1.0\columnwidth]{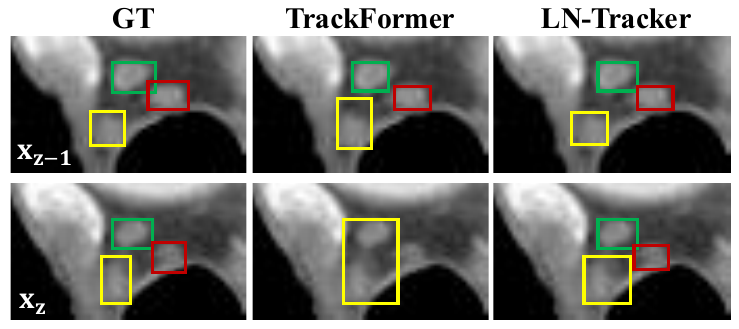}
\caption{Detection and tracking results on two consecutive slices. Previous MOT method, TrackFormer~\cite{meinhardt2022trackformer}, makes a failed large box prediction (\textcolor{yellow}{yellow box}) and loses the tracking of other two LNs (\textcolor{green}{green box} and \textcolor{red}{red box}).  }
\label{fig:trackformer}
\vspace{-2mm}
\end{figure}

In this work, we propose a novel end-to-end trainable LN tracking transformer, named LN-Tracker, to achieve joint detection and 3D instance association (merging). A recently developed strong 2.5D DETR-based LN detector (LN-DETR)~\cite{yu2025effective} is adopted as our detection backbone. To initiate the tracking ability of LN instances, we follow the transformer-based multi-object tracking (MOT) setup of~\cite{meinhardt2022trackformer,zeng2022motr}. Specifically, LN-Tracker first decouples transformer decoder's query into the track and detection groups, where track queries autoregressively follow previously tracked LN instances along the z-axis of a CT scan. Track queries are used to follow existing LN instances, dynamically updating with varying lengths at each slice, while detection queries are employed to identify newly appearing LNs. In LN detection scenario, tracking the instance association between slices is more challenging than that in natural video tracking (targets such as persons typically exhibit distinct visual features that are easier to detect and associate). Because LNs possess extremely similar intensity, size and shape to adjacent tissues. Hence, previous MOT models struggle to detect and associate correct 3D LN instances, as shown in Figure \ref{fig:trackformer}.  To address this, we introduce a similarity loss that encourages the network to predict higher similarity scores for the same LN across consecutive slices, enhancing tracking accuracy in this low-contrast context. Additionally, in original MOT configuration~\cite{meinhardt2022trackformer,zeng2022motr}, detection query interacts with track query during the self-attention operation in decoder, which is observed to negatively impact detection query's accuracy. Specifically, when a track query and a detection query predict the same LN, the track query has higher matching priority for the instance identity propagation, which causes the detection query to be falsely penalized under DETR’s one-to-one matching scheme during training, decreasing the detection query's accuracy. To solve this, we design a new masked attention module in the transformer decoder to prevent detection query from accessing track query in self-attention operation, which preserves the independence of detection query, improving overall 3D detection and tracking accuracy.

Our main contributions can be summarized as follows:
\begin{itemize}
\item To the best of our knowledge, we are the first to tackle 3D lesion detection from a tracking perspective, demonstrating its effectiveness across various 3D volumetric detection tasks.

\item We propose an end-to-end trainable LN tracking transformer, LN-Tracker, to achieve joint detection and 3D LN instance association. With a new masked attention module and an inter-slice similarity loss, LN-Tracker preserves detection query's high accuracy and encourages the cohesive LN association between consecutive slices. 



\item We conduct extensive evaluation on 4 LN CT datasets ($\ge2.7\%$ gain in average sensitivity when compared to 3D and 2.5D detectors), and further validate the generalizability on various targets (lung nodules and prostate tumor) and modality (MRI) by achieving top performance.


%

\end{itemize}

\section{Related work}
\label{sec:related}

\subsection{Object detection in medical images}
Lesion detection in medical images involves to localize the lesion from an image via regressing a set of Bbox coordinates with the associated foreground probability predictions. 
For 2D detection, previous studies usually leverage detectors designed for natural images to adapt to medical tasks, such as detection of skin lesion~\cite{nida2019melanoma}, traumatic injury~\cite{ju2023fracture}, and mammography mass~\cite{yang2021momminet}. For 3D detection, both 3D detectors and 2.5D detectors are the common choice. 3D detectors explicitly model the 3D spatial context via 3D conv kernels more suitable for application in isotropic volumetric images~\cite{wang2020focalmix,baumgartner2021nndetection,brugnara2023deep}. As comparison, the 2.5D detector uses a 2D network with multiple input slices~\cite{yan2019mulan,yan2020learning,yang2021asymmetric,yu2025effective}, which leverages the pre-trained 2D model weights and often shows improved accuracy as compared to 3D detectors in anisotropic CT scans.


\subsection{LN detection and segmentation}
Automatic LN detection has been exploited by mainly focusing on extracting effective LN features, incorporating organ priors or utilizing advanced learning models. Considering the small nature and the usually anisotropic spacing in diagnostic CT scans (z-spacing$\gg$x,y-spacing), 2.5D CNN detectors are mostly applied. For example, a 2.5D CNN-based detector incorporated with LN station information~\cite{guo2021deepstationing} is proposed to detect the mediastinal LNs~\cite{yan2023anatomy}. A transformer-based 2.5D detector, LN-DETR~\cite{yu2025effective}, is developed upon Mask-DINO~\cite{li2023mask} to detect LNs in several body regions such as neck, mediastinal and upper abdomen. Another line of works uses a 3D CNN-based segmentation model for LN detection~\cite{zhu2020lymph,zhu2020detecting,bouget2023mediastinal,guo2022thoracic}; however, 
voxel-wise segmentation loss is difficult to supervise the instance-oriented target learning.  In this work, we tackle 3D LN detection from tracking perspective and show its effectiveness in four LN datasets of various body regions and diseases.


\subsection{Multi-Object Tracking}
Recent progress in MOT relies on a "tracking-by-detection" approach, where objects are first detected in each frame and then linked across frames, usually by using Kalman filter or appearance re-identification features. DeepSORT~\cite{wojke2017simple}, ByteTrack~\cite{zhang2022bytetrack}, and FairMOT~\cite{zhang2021fairmot} inherit this paradigm with different association design. While ByteTrack solely relies on a two-stage matching strategy with the motion prior derived from Kalman filter, DeepSORT and FairMOT compute a deep association metric for re-identification. Recently, Transformer-based approaches like TransTrack~\cite{sun2020transtrack}, TrackFormer~\cite{meinhardt2022trackformer} and MOTR~\cite{zeng2022motr} have brought a shift in the tracking-by-detection paradigm by using self-attention mechanism to effectively model targets temporal dependency between frames. TrackFormer~\cite{meinhardt2022trackformer} and MOTR~\cite{zeng2022motr}, for instance, introduce the track query and generate object embeddings directly from a transformer-based attention mechanism, which dynamically links objects across frames without requiring the explicit re-identification module. In LN tracking, due to LN's extremely similar intensity, size and shape with adjacent tissues, TrackFormer or MOTR struggles to detect and associate correct 3D LN instances. We introduce a similarity loss supervised on track queries and a new masked attention module in the transformer decoder to solve this issue.

\section{Methods}
\label{sec:methods}

\begin{figure*}[t]
   \begin{center}
      \includegraphics[width=1.0\linewidth]{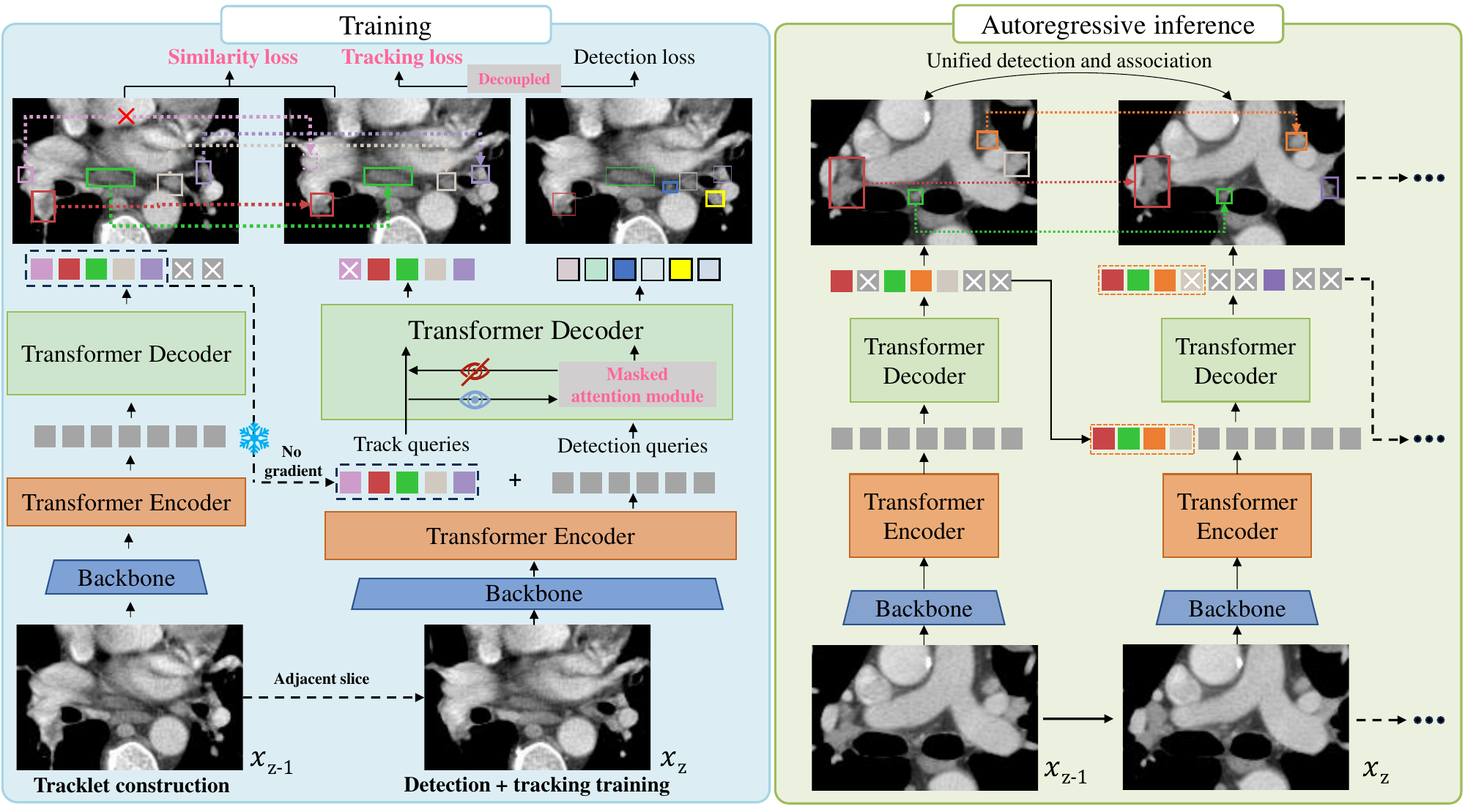}
   \end{center}
   \caption{Overall training and inference framework of the proposed LN-Tracker, where LN instance information from previous slice can be propagated to the current slice, promoting detection cohesion and consistency without involvement of offline post-processing. A pink color indicates the technical contribution of the proposed LN-Tracker. Note that in the self-attention computation of transformer decoder, track queries have access to detection queries for the context alignment, while detection queries are blocked from accessing track queries. }
   \label{fig:framework}
   \vspace{-4mm}
\end{figure*}

In this section, we introduce LN-Tracker, the unified framework for LN detection and tracking in 3D CT scans, which is built upon the detection-transformer architecture. Sec.~\ref{subsec:formulation} first describes the proposed 3D detection by tracking formulation. In Sec.~\ref{subsec:track_queries}, we elaborate on the training procedure and introduce the masked attention module and the similarity loss to improve the association of 3D LNs and achieve improved 3D detection accuracy. Finally, Sec.~\ref{subsec:infer} outlines the autoregressive inference procedure of LN-Tracker.

\subsection{3D LN detection as a tracking task}
\label{subsec:formulation}

As discussed previously, pure 3D detector suffers from the training challenges and 2.5D approach is prone to produce inconsistent predictions due to the lack of inter-slice interaction in training and the heuristic post-merging in inference. To conquer these difficulty, a potential way is to leverage pretrained 2.5D detectors to address the challenges of 3D network training and meanwhile incorporating explicit spatial-temporal constraints to facilitate the cohesive and consistent 3D prediction. This inspires us to formulate 3D LN detection as a multi-object tracking task, aiming to solve the difficulties of 3D training without sacrificing the consistency and accuracy of producing 3D predictions. Figure~\ref{fig:framework} illustrates the overview of LN-Tracker framework. 

Built upon a recent state-of-the-art LN detection transformer, LN-DETR~\cite{yu2025effective}, our LN-Tracker consists of three base components: (1) a 2.5D feature extraction backbone that generates and fuses image features via a CNN network (\eg, ResNet-50), (2) a transformer encoder that takes the flattened CNN features with positional encodings as input and produces encoded queries and initial anchor predictions, and (3) a transformer decoder that takes the selected queries from encoder with the associated anchor coordinates as input and outputs the final detection results. LN-Tracker tracks LN instances along the z-axis in a CT scan. To initiate the tracking ability, we follow TrackFormer~\cite{meinhardt2022trackformer}, which introduces track queries to allow simultaneous detection and temporal aggregation. Here, track queries are used to follow existing LN instances, dynamically updating with varying lengths at each slice, and detection queries are employed to identify newly appearing LNs.

\subsection{Unified LN detection and tracking}
\label{subsec:track_queries}

\noindent
\textbf{Track query initialization.} Figure~\ref{fig:framework} demonstrates the training procedure of the proposed LN-Tracker. Given a pair of adjacent slices $x_{z-1}$ and $x_z$ sampled during training, the detector first generates LN box predictions for the reference slice $x_{z-1}$. Hungarian matching is then performed to obtain queries that have box predictions matched with the ground truth boxes in $x_{z-1}$. The matched queries are then initialized as the track queries for training slice $x_z$.



\noindent
\noindent
\textbf{Decoupled training.} In the default setup of TrackFormer~\cite{meinhardt2022trackformer}, track queries and detection queries are concatenated and passed into the decoder for simultaneous prediction and joint training. Following the DETR scheme, one-to-one matching is used to assist in the calculation of the detection loss among predictions and ground truth boxes. To fulfill unified detection and tracking, track queries have higher matching priority than detection queries to maintain the tracked instances. However, this scheme encounters a significant performance drop in LN detection. Different from natural video tracking (targets typically exhibit distinct visual features that are easier to detect and associate), LN detection is more complex as they tend to cluster closely with low contrast to their surrounding tissues, which also have similar shapes and sizes. Given this challenge, we observed that track queries could sometimes be mistakenly matched to a nearby new detection, causing detection queries at the same location to be paradoxically penalized, leading to degraded detection capacity. To maintain the high detection capacity of the detection query, we employ a decoupled training strategy that explicitly decouples the training of track queries and detection queries so that the detection component can produce a complete set of predictions independent of tracking status (see the decoupled loss in Figure~\ref{fig:framework}). Specifically, for tracking, we jointly apply an instance affinity classification loss (where 1 denotes the same instance and 0 denotes different instances) alongside the associated box regression loss to guide inter-slice association. The tracking loss is defined as:
\begin{equation}
    \mathcal{L}_{track} = \lambda_{cls} \mathcal{L}_{cls}^{track} + \lambda_{box} \mathcal{L}_{box}^{track}
\end{equation}
where $\mathcal{L}_{cls}$ is the focal loss, and $\mathcal{L}_{box}$ is a combination of L1 loss and GIoU loss.

For detection, since it has been decoupled from tracking, we follow the widely adopted DETR training loss by calculating the pairwise matching cost between detector predictions and all ground truth boxes within a slice. The detection loss is defined as:
\begin{equation}
    \mathcal{L}_{det} = \lambda_{cls} \mathcal{L}_{cls}^{det} + \lambda_{box} \mathcal{L}_{box}^{det}
\end{equation}

\begin{figure}[htb]
   \begin{center}
      \includegraphics[width=.95\linewidth]{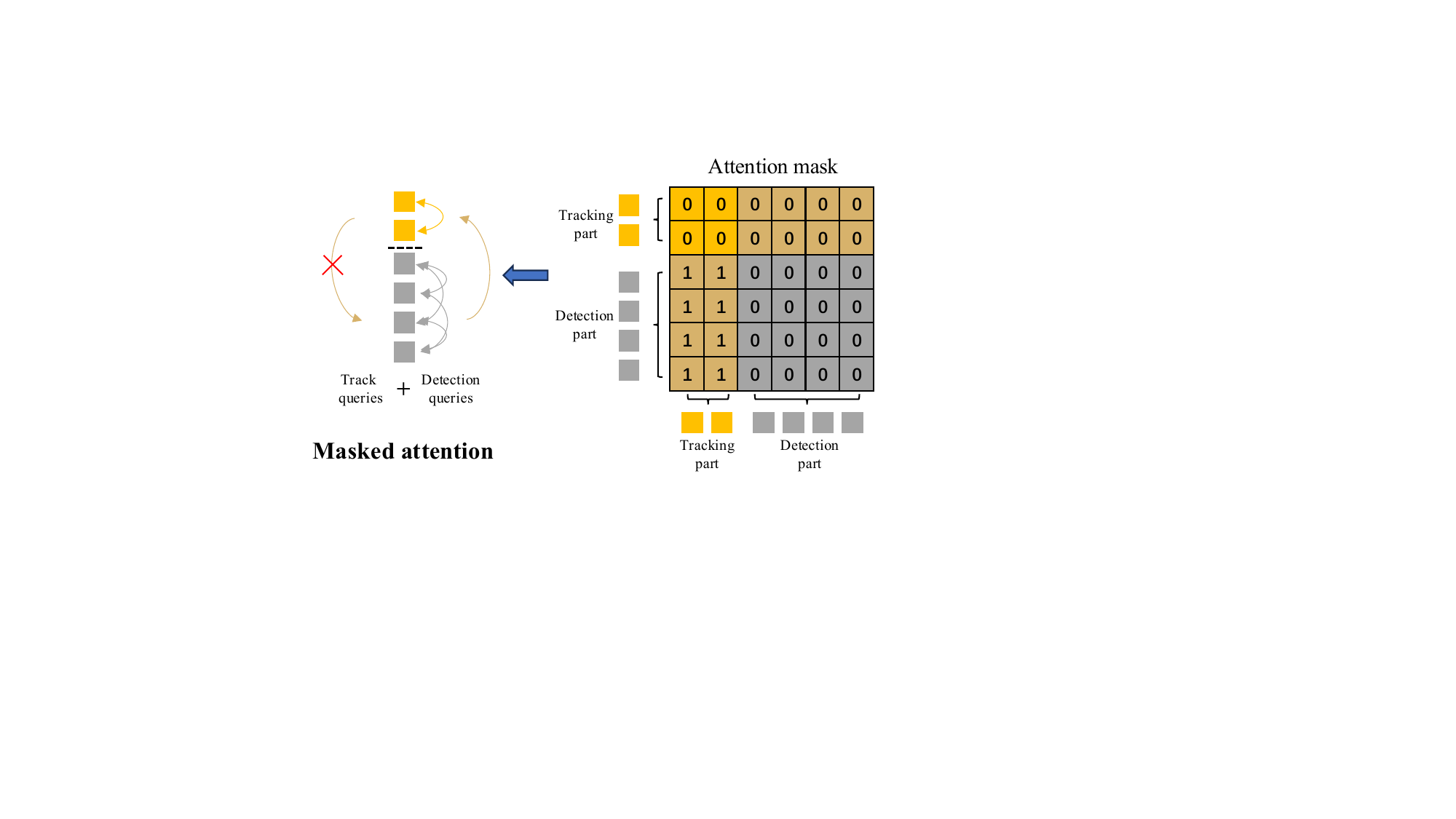}
   \end{center}
   \caption{Illustration of the masked attention module between track and detection queries. The elements 1 in attention mask mean the detection queries cannot see the track queries. }
   \label{fig:atten_mask}
   \vspace{-2mm}
\end{figure}

\noindent
\textbf{Masked attention.}
In the self-attention mechanism of the transformer decoder, the interaction between track queries and detection queries allows the location priors carried by track queries to influence the detection queries in the current slice.  This may encourage the detection queries to learn a shortcut during decoding, resulting in their degraded sensitivity (as demonstrated in Table~\ref{tab:ablation_study}). To address this issue, we introduce a masked attention module in the transformer decoder. This module allows track queries to access detection queries for context alignment within the current slice while blocking information leakage from track queries, thereby preventing detection queries from being influenced by track queries. To achieve this, we add an attention mask denoted as $\textbf{M}=[m_{ij}]_{N\times N}$, in transformer decoder, where $N = N_{track}+N_{det}$. $N_{track}$ and $N_{det}$ are the number of track and detection queries as in Figure \ref{fig:atten_mask}. We let the first $N_{track} \times N_{track}$ rows and columns represent the tracking part, and the latter represents the detection part. $m_{ij}=1$ means the $i$-th query cannot see the $j$-th query and $m_{ij}=0$ otherwise. The attention mask is then devised as follows:
\begin{equation}
m_{ij}= \begin{cases} 1, & \text { if } j<N_{track} \text { and } i \geq N_{track} \\ 0, & \text { otherwise. }\end{cases}
\end{equation}

\noindent
\textbf{Enhanced matching via similarity loss.} As shown in Figure~\ref{Fig:track_motivation} and \ref{fig:trackformer}, LNs tend to cluster closely with indistinct visual features, posing a challenge for inter-slice association. To address this, we introduce a similarity loss $\mathcal{L}_{sim}$ that encourages track queries belonging to the same instance across slices to be close in the latent space, while those from different instances are pushed apart. Specifically, given a pair of reference track queries $\mathcal{Q}_{z-1}^{track} \in \mathbb{R}^{N_{track} \times C}$ from adjacent slice $x_{z-1}$ and the updated track queries $\mathcal{Q}_{z}^{track}$ from the current slice $x_z$, an MLP layer $\phi$ first projects these queries into a new latent space. Then, similarity logits are generated using pairwise dot-products, which are then normalized by a sigmoid activation to compute binary cross-entropy loss against the affinity matrix $\textbf{A}\in\mathbb{R}^{N_{track} \times N_{track}}$, where 1 represents the same instance and 0 represents different instances (Supplementary Figure 1). 
\begin{equation}
    \mathcal{L}_{sim} = \mathcal{L}_{bce}(\text{sigmoid}(\phi(\mathcal{Q}_{z-1}^{track})\phi(\mathcal{Q}_{z}^{track})^T),\textbf{A})
\end{equation}
We also use this similarity measure as an auxiliary criterion for the instance association during inference.

Finally, the total training loss is the sum of $\mathcal{L}_{track}$ , $\mathcal{L}_{det}$, and $\mathcal{L}_{sim}$ .




\subsection{LN-Tracker inference}
\label{subsec:infer}
During the inference stage, detection begins from the bottom slice of the CT scan. When a slice yields valid object detections with classification scores above the object score threshold, these detected boxes are initialized as tracklets, and the corresponding track queries are propagated to the next slice. In each subsequent slice, the number of track queries dynamically adjusts as new objects are detected or existing tracklets are removed. We conduct this autoregressive process until the last slice. Finally, we obtain the 3D detection boxes from the valid tracklets without further post-processing. The pseudo-code for LN-Tracker inference is provided in Algorithm \ref{algo:pseudo}.

\begin{algorithm}[!h]
\small
\SetAlgoLined
\DontPrintSemicolon
\SetNoFillComment
\KwIn{CT scan $\mathcal{X}$; object detector $\texttt{Det}$; detection score threshold $\tau_{det}$; tracking score threshold $\tau_{track}$; ReID similarity score threshold $\tau_{sim}$}
\KwOut{3D LN bounding boxes $B$}

Initialization: $B, \mathcal{T}, \mathcal{Q}_0^{track} \leftarrow \emptyset,\emptyset,\emptyset$\;
\For{$x_z$ \textnormal{in} $\mathcal{X}$}{
$\mathcal{Q}_z^{track},\mathcal{Q}_z^{det},\mathcal{D}_z^{track},\mathcal{D}_z^{det}\leftarrow \texttt{Det}(x_z,\mathcal{Q}_{z-1}^{track})$ \;
        \tcc{Hard assign identity \& update tracklet}
	\For{$q',q,d$ \textnormal{in zip}$(\mathcal{Q}_{z-1}^{track},\mathcal{Q}_z^{track},\mathcal{D}_z^{track})$}{
        $k = q'.id$\;
	\If{$d.score > \tau_{track}~\textnormal{and}~sim(q',q)>\tau_{sim}$}{
        $\mathcal{T}_k \leftarrow \mathcal{T}_k \cup \{d\}$\;
	}
	\Else{
          $B \leftarrow B \cup \{\text{3D box from}~\mathcal{T}_k\}$ \;
          $\mathcal{T} \leftarrow \mathcal{T} \setminus \{\mathcal{T}_k \}$\;
          $\mathcal{Q}_z^{track} \leftarrow \mathcal{Q}_z^{track} \setminus \{q\}$ \;
	}}
        \tcc{delete re-detection objects \& add new born objects}
        \For{$q,d$ \textnormal{in zip}$(\mathcal{Q}_z^{det},\mathcal{D}_z^{det})$}{
	\If{$d.score > \tau_{det}$}{
            Associate $\mathcal{T}$ and $d$, get match id $k$\;
            \If{$k==-1$}{
            $\mathcal{T} \leftarrow \mathcal{T} \cup \{d\}_{new-id}$\;
            $\mathcal{Q}_z^{track} \leftarrow \mathcal{Q}_z^{track} \cup \{q\}$\;
            }
	}
	}
}
Return: $B$
\caption{Pseudo-code of LN-Tracker.}
\label{algo:pseudo}
\end{algorithm}

\section{Experiments}
\label{sec:exps}
In this section, we first evaluate our LN-Tracker in LN detection task, and four LN datasets (1 public and 3 private) of different body regions and diseases are collected with a total of 1,000 patients. To further demonstrate the effectiveness and generalizability of LN-Tracker, we report quantitative results on two public 3D lesion detection tasks. Finally, we verify contributions of individual component of LN-Tracker in an ablation study. Results from state-of-the-art 3D, 2.5D lesion detection and MOT methods are also compared.

\subsection{Experimental setup}
\noindent
\textbf{LN Datasets.} For the LN detection task,  we collect 4 LN datasets of different body regions (neck, chest, and abdomen) and various diseases (head \& neck cancer, lung cancer, esophageal cancer, and pancreatic cancer) containing a total of 1,000 patients and 7252 3D annotated LN instances. Among them, NIH-LN~\cite{bouget2019semantic} is a public LN dataset, while the rest are from three clinical centers (denoted as HN-LN, Eso-LN, and Pan-LN for simplification). The details of the dataset can be seen in Table \ref{tab:dataset}. We randomly split each dataset into 70\% for training, 10\% for validation, and 20\% for testing at the patient level.

\noindent
\textbf{Other lesion detection datasets.} We further evaluate our method on two public medical detection tasks: lung nodule detection and prostate tumor detection. Public LIDC-IDRI~\cite{armato2011lung} dataset is used for lung nodule detection consisting of 1012 chest CT scans with manual 3D lung nodule segmentation. Prostate tumor detection is conducted on prostateX~\cite{cuocolo2021quality,litjens2014computer} dataset, which has 200 labeled T2/ADC MRI sequences (only T2 modality is used in our experiment as multi-modality learning is not the focus of this work). Similar to LN detection task, we also randomly split each dataset at 70\%, 10\%, and 20\% for training, validation and testing.

\noindent
\textbf{Evaluation metrics.} We define that given predictions and ground truths, an object is considered detected if the intersection over union (IoU) between any detected 3D boxes and its GT 3D box is greater than 0.1. Following previous works \cite{yu2025effective,jaeger2020retina,isensee2021nnu}, we report the average sensitivity (AS) over 1, 2, 4, and 8 FPs per patient and average precision (AP) at 0.1 IoU threshold in 3D.

\begin{table}[h]
    \centering
    \caption{Statistics of 4 LN detection datasets. Each dataset is randomly split at 70\%, 10\%, and 20\% for training, validation, and testing.  }
    \resizebox{0.48\textwidth}{!}{%
        \setlength\tabcolsep{3.0pt}
        \scalebox{0.98}{
    \begin{tabular}{l|c|c|c|c}
    \toprule
        Dataset & \#Patients & \#LNs & Avg. Res. (mm)  & Cancer Type \\ \hline
        NIH-LN & 89 & 1956 &  $\left(0.82, 0.82, 2.0\right)$ & lung cancer\\
        HN-LN & 256 & 1890 & $\left(0.46,0.46,4.0\right)$ &head \& neck cancer  \\
        Eso-LN & 300 & 2515 & $\left(0.70,0.70,4.9\right)$ & esophageal cancer \\
        Pan-LN& 355 & 1178 & $\left(0.68, 0.68, 0.8\right)$ &pancreatic cancer \\ 
    \bottomrule
    \end{tabular}
    }}
    \label{tab:dataset}
\end{table}

\noindent
\textbf{Comparing methods.} We conduct extensive comparison evaluation for all detection tasks, including 3D segmentation method nnUNet~\cite{isensee2021nnu}, 3D detection method nnDetection~\cite{baumgartner2021nndetection}, 2.5D CNN-based lesion detector MULAN~\cite{yan2019mulan}, transformer-based LN detector LN-DETR~\cite{yu2025effective}, instance segmentation model Mask2Former~\cite{cheng2022masked}, VITA~\cite{heo2022vita}, and GenVIS~\cite{heo2023generalized}, MOT model TrackFormer~\cite{meinhardt2022trackformer}, and latest prompt-based video segmentation model SAM2~\cite{ravi2024sam2segmentimages} with LN-DETR's prediction as prompts. Note that except for 3D methods, all other models are equipped with the same 2.5D backbone as our LN-Tracker to ensure a fair comparison.

\begin{table*}[h]
\renewcommand\arraystretch{1.2}
    \centering
    \caption{Results for the proposed and comparison detection methods on four LN datasets. Note that nnUNet is a 3D segmentor thus only has one FP point and no AP metric. Best in \textbf{bold}, second \underline{underline}. \textit{AS: average sensitivity over 1, 2, 4, and 8 FPs per patient. AP: average precision at 0.1 IoU threshold in 3D.}}
    \setlength\tabcolsep{3.0pt}
    \resizebox{0.87\width}{!}{
    \begin{tabular}{lcccc|cc|cc|cc|cc}
        \toprule
        \multirow{2}{*}{Methods} & \multirow{2}{*}{Types~~} & \multirow{2}{*}{GFLOPs~~} & \multicolumn{2}{c|}{NIH-LN} & \multicolumn{2}{c|}{HN-LN} & \multicolumn{2}{c|}{Eso-LN} & \multicolumn{2}{c|}{Pan-LN} & \multicolumn{2}{c}{Mean} \\ 
        \cline{4-13}
        & & & AS $\uparrow$ & $\text{AP}_{10}^{3D}\uparrow$ & AS $\uparrow$ & $\text{AP}_{10}^{3D}\uparrow$ & AS $\uparrow$ & $\text{AP}_{10}^{3D}\uparrow$ & AS $\uparrow$ & $\text{AP}_{10}^{3D}\uparrow$  & AS $\uparrow$ & $\text{AP}_{10}^{3D}\uparrow$ \\ 
        \midrule
        nnUNet~\cite{isensee2021nnu}  &\multirow{2}{*}{3D}  & 735.38 & \multicolumn{2}{c|}{56.17@6.29FPs}   & \multicolumn{2}{c|}{55.48@2.86FPs} & \multicolumn{2}{c|}{61.32@2.17FPs}  & \multicolumn{2}{c|}{33.51@1.18FPs}  &-- &--         \\ 
        nnDetection~\cite{baumgartner2021nndetection}   &  & 814.74 &51.49    &\underline{50.66}     &64.55    &48.30    &66.32     &59.32    &\underline{52.45}     &\underline{38.78}     &58.72    &49.34      \\
        \midrule
        Mask2Former~\cite{cheng2022masked} &\multirow{5}{*}{2.5D} & 60.78 & 50.00  & 45.43  & 60.45  & 46.47 & 62.20 & 51.24  & 38.92  &21.27  & 52.89  &41.10               \\
        VITA~\cite{heo2022vita}  &  & 105.39 & 50.21    & 45.10   & 59.25   & 46.37  & 68.29   & 55.36   & 40.21     &24.06  &54.49  &42.72              \\
        GenVIS~\cite{heo2023generalized}  &  &110.64 & 51.80    & 47.22   & 62.64   & 47.12  & 67.55   & 54.53   & 45.09     & 29.40 & 56.77  &44.56              \\
        MULAN~\cite{yan2019mulan}&  & 102.11 & 52.98  & 48.12 & \underline{66.27}  & \underline{51.38} & 67.94 & 58.75 & 45.88 & 32.06  &58.27 &47.58             \\
        LN-DETR~\cite{yu2025effective}& & 141.08 & \underline{54.04}  & 50.08  & 64.90& 50.78  & \underline{69.86}  & \underline{62.53}  & 51.03  & 37.85  &\underline{59.96}  & \underline{50.31}              \\
        \midrule
        TrackFormer~\cite{meinhardt2022trackformer} &\multirow{3}{*}{Tracking} & 130.66 & 29.68  & 23.35  & 48.29  & 36.93  & 56.27  & 49.79  & 47.94  &33.38                &45.55  &35.86               \\
        LN-DETR+SAM2~\cite{ravi2024sam2segmentimages} & & -- &45.87 &43.78 &58.87 &43.86 &55.83 & 47.28 &40.30 &25.78 &50.22 &40.18 \\
        \cellcolor{gray!20}LN-Tracker (Ours) & & \cellcolor{gray!20}142.10 & \cellcolor{gray!20}\textbf{57.66}   & \cellcolor{gray!20}\textbf{54.33}   & \cellcolor{gray!20}\textbf{66.95}   & \cellcolor{gray!20}\textbf{53.19}  & \cellcolor{gray!20}\textbf{69.95}   & \cellcolor{gray!20}\textbf{63.12}   & \cellcolor{gray!20}\textbf{56.06}                & \cellcolor{gray!20}\textbf{40.06}  &\cellcolor{gray!20}\textbf{62.66}  &\cellcolor{gray!20}\textbf{52.68}   \\

        \bottomrule
    \end{tabular}
    }
    \label{tab:LN_detection}
\end{table*}

\subsection{Implementation details}
Following \cite{yu2025effective}, LN-Tracker adopts ResNet50 with the 2.5D feature fusion layer as CNN backbone and the leading DETR-based object detection method Mask DINO \cite{li2023mask} as the transformer encoder-decoder architecture. In training, each mini-batch consists of 8 pairs of neighbor slices. By the nature of track queries, each slice may have a different number of total queries $N = N_{det} + N_{track}$. In order to do parallel training, we pad track queries with additional zero-initialized queries and set fixed 300 detection queries for each slice.  We use RAdam optimizer with the initial learning rate of $2 \times 10^{-4}$ and a weight decay of $1 \times 10^{-4}$ for 30 epochs training. Besides, cosine annealing scheduler is adopted to reduce the learning rate to $1 \times 10^{-5}$ with a warm-up step of 500 iterations. During inference, the queries with corresponding classification scores above 0.3 (\ie, $\tau_{det}$ and $\tau_{track})$ will be regarded as the LN predictions. And the ReID similarity score threshold $\tau_{sim}$ is set to 0.5 to decide whether the track query should keep the same identity between slices. For more detailed hyper-parameters and dataset pre-processing operations, we refer to the appendix.

\subsection{Experimental results}
\noindent
\textbf{LN detection results.} The quantitative evaluation on four LN detection datasets is presented in Table~\ref{tab:LN_detection}. As shown, our LN-Tracker surpasses all compared methods across the four datasets, outperforming the closest competitor, LN-DETR (the state-of-the-art 2.5D LN detector), by $2.7\%$ in mean AS ($62.66\%$ vs. $59.96\%$) and $2.37\%$ in mean AP ($52.68\%$ vs. $50.31\%$).
Notably, for HN-LN and Eso-LN datasets (with thicker image spacing in z-axis as shown in Table \ref{tab:dataset}), the 2.5D methods (\eg, MULAN and LN-DETR) generally outperform the 3D detector nnDetection. This indicates when adjacent slices have higher variations in the context, inter-slice modeling using 3D kernels becomes ineffective. Conversely, on the Pan-LN dataset with thinner slices, nnDetection performs better than all 2.5D methods, though still lower than the performance of our LN-Tracker, demonstrating the effectiveness and generalization of our tracking-based method on both thick and thin z-axis spacing configurations.
On the other hand, instance segmentation approaches, such as VITA and Mask2Former, generally perform inferior when compared to 2.5D and 3D detection methods, with noticeable gaps of at least $3.78\%$ and $4.86\%$ in mean AS and AP, respectively. nnUNet, as a detection-by-segmentation model, does not differentiate LN instances generally producing less favorite results. TrackFormer performs the worst across three tasks, highlighting the limitations of natural video tracking approaches when directly applied to the volumetric lesion detection task in medical imaging. When combining fully trained LN-DETR and SAM2 like Grounded-SAM2\footnote{\url{https://github.com/IDEA-Research/Grounded-SAM-2}} leads to markedly inferior performance than the standalone LN-DETR.


\begin{figure*}[t]
   \begin{center}
      \includegraphics[width=.96\linewidth]{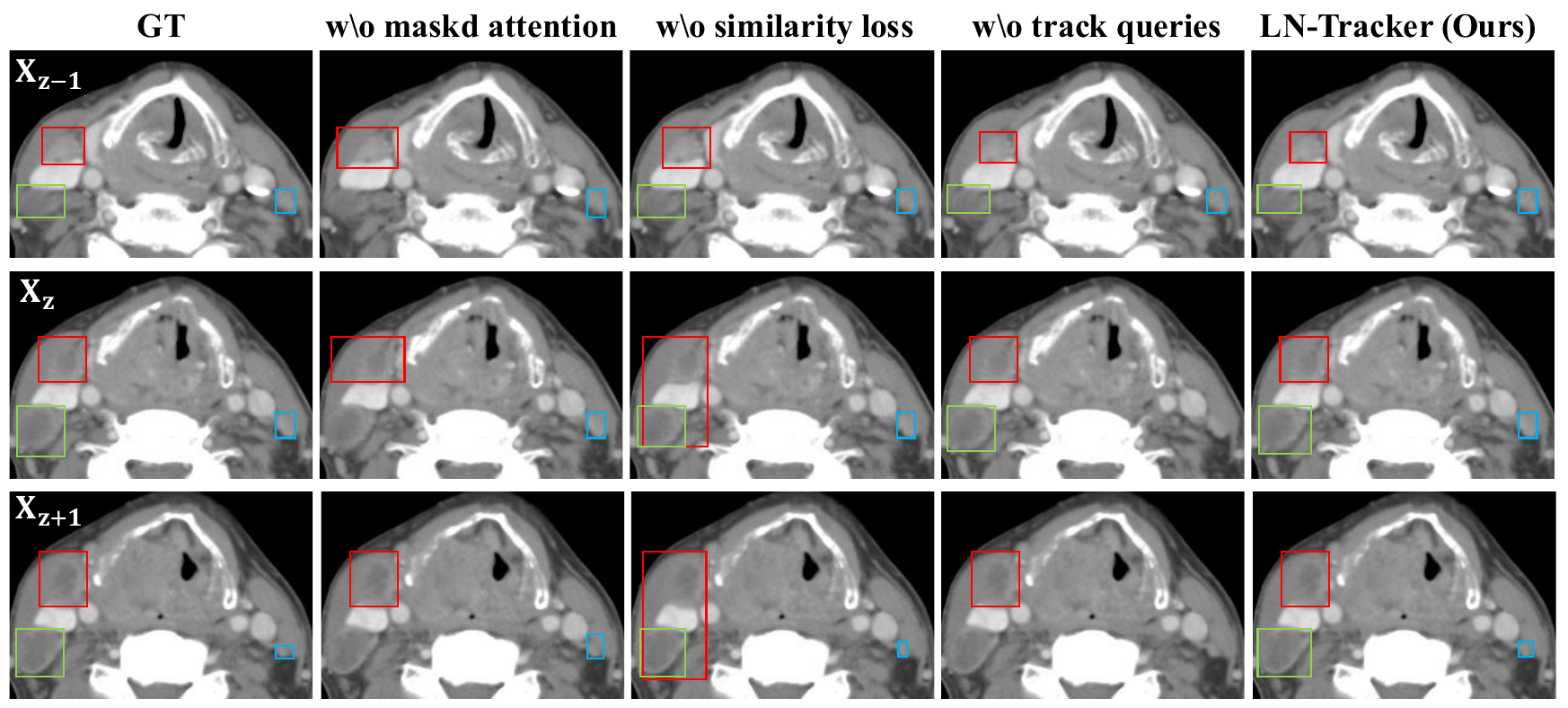}
   \end{center}
   \caption{Qualitative detection results across consecutive CT slices. In the GT column, same colored boxes denote the same LN instance across slices, with missing boxes indicating missed detections. In the 2nd column, without masked attention module, it decreases the sensitivity of our LN-Tracker. In the 3rd column, removing the similarity loss causes the tracking queries to incorrectly make predictions involving adjacent LNs. In the 4th column, excluding tracking queries during inference results in inconsistent LN predictions across slices.}
   \label{fig:qualitative}
\end{figure*}

\begin{table}[h]
\renewcommand\arraystretch{1.2}
    \centering
    \caption{Comparsions of lung nodule detection and prostate tumor detection performance on the \textbf{public} LIDC-IDRI and ProstateX datasets. }
    \setlength\tabcolsep{2.0pt}
    \resizebox{0.8\columnwidth}{!}{
    \begin{tabular}{lcc|cc}
        \toprule
        \multirow{2}{*}{Methods}  & \multicolumn{2}{c|}{LIDC-IDRI}& \multicolumn{2}{c}{ProstateX}\\ 
        \cline{2-5}
        & AS $\uparrow$ & $\text{AP}_{10}^{3D}\uparrow$ & AS $\uparrow$ & $\text{AP}_{10}^{3D}\uparrow$ \\ 
        \midrule
        nnUNet~\cite{isensee2021nnu}   & \multicolumn{2}{c|}{39.86@0.15FPs} & \multicolumn{2}{c}{7.55@0.31FPs}  \\ 
        nnDetection~\cite{baumgartner2021nndetection}     &74.76     &66.18  &33.96 &20.44\\
        \midrule
        Mask2Former~\cite{cheng2022masked}  & 60.78  & 52.34  &\underline{44.78} &23.90 \\
        VITA~\cite{heo2022vita}    & 70.92    & 59.84 &41.95 &22.55  \\  
        MULAN~\cite{yan2019mulan}      & 74.28  & 62.96 &25.85 &17.68 \\
        LN-DETR~\cite{yu2025effective}  & \underline{76.54}  & \underline{67.13} &38.56 &\underline{24.71} \\
        \midrule
        TrackFormer~\cite{meinhardt2022trackformer}  & 65.79  & 55.23 &30.24 &20.25  \\
        \cellcolor{gray!20}LN-Tracker  & \cellcolor{gray!20}\textbf{77.76}   & \cellcolor{gray!20}\textbf{68.46}  & \cellcolor{gray!20}\textbf{48.73}   & \cellcolor{gray!20}\textbf{29.50}   \\

        \bottomrule
    \end{tabular}
    }
    \label{tab:lidc}
    \vspace{-2mm}
\end{table}

Regarding the GFLOPs, it is observed that 2.5D and tracking-based LN methods are much more efficient than the 3D methods. Specifically, our method accounts only 17\% GFLOPs of the 3D detector nnDetection, and it has a neglecting increasing ($<$1\%) of GFLOPs, yet with 2.7\% AS improvement, when compared to the 2nd best performing 2.5D detector LN-DETR.

\noindent
\textbf{Evaluation on the public LIDC-IDRI and ProstateX dataset.} Detailed results on the public lung nodule and prostate tumor detection tasks are shown in Table \ref{tab:lidc}. It can be seen that LN-Tracker achieves the highest AS and AP among all compared methods on both tasks. On LIDC dataset, LN-Tracker surpasses the second-highest LN-DETR by $1.22\%$ in AS and $1.33\%$ in AP. Same as LN detection, 2.5D and 3D detectors (LN-DETR, nnDetection, and MULAN) substantially outperform the instance segmentation approaches (VITA and Mask2Former) in lung nodule detection. 
On ProstateX dataset, LN-Tracker also achieves the top performance with larger accuracy gap as compared to other competitors, e.g., outperforming the closest Mask2Former by $3.95\%$ in AS and $5.60\%$ in AP. Note that ProstateX is an MRI dataset and lesions in T2 scans often exhibit subtle intensity changes. By demonstrating stronger performance on the challenging and cross-modality ProstateX dataset, we further validate the effectiveness and generalizability of the proposed method.

\noindent
\textbf{Ablation results.} We conduct the ablation study using the public NIH-LN dataset to demonstrate the effectiveness of the proposed components by removing one element at a time and assessing the resulting performance change. The quantitative and qualitative results are shown in Table \ref{tab:ablation_study} and Figure \ref{fig:qualitative}, respectively. First, we remove the masked attention module in LN-Tracker, leading to a significant reduction of $7.87\%$ AS and $14.45\%$ in AP, which highlights the importance of preventing the information leakage from track queries to detection queries for achieving robust and cohesive detection and association. Then, by removing the similarity loss, which implicitly promotes the association of the same instance across slices, we observe a decrease of $2.87\%$ in AS and $1.08\%$ in AP. Finally, we demonstrate the superiority of unified detection and tracking using track queries. Adopting the same training configuration as LN-Tracker but removing track queries at inference time, we observe a performance drop of $4.57\%$ in AS and $3.88\%$ in AP, demonstrating the effectiveness of track queries for temporal aggregation.

\begin{table}
    \begin{center}
        \caption{
        Ablation study for LN-Tracker components on NIH-LN~\cite{bouget2019semantic}. The last row without track queries means only the predictions of detection queries are used to merge tracks as in \cite{yan2020learning}.
    }
        \setlength\tabcolsep{2.0pt}
        \begin{tabular}[t]{lcccc}
            \toprule
            Method      & \multicolumn{1}{c}{AS $\uparrow$} & $\Delta$ & \multicolumn{1}{c}{$\text{AP}_{10}^{3D} \uparrow$} & $\Delta$\\

            \midrule
            \cellcolor{gray!20}LN-Tracker      & \cellcolor{gray!20}57.66 &\cellcolor{gray!20}- & \cellcolor{gray!20}54.33 &\cellcolor{gray!20}- \\
            -------w\textbackslash o--------& \multicolumn{4}{c}{-----------------------------------------}\\
            Masked attention      &49.79  &\textcolor{blue}{-7.87}  &39.88  &\textcolor{blue}{-14.45} \\
            Similarity loss     &54.79  &\textcolor{blue}{-2.87}  &53.25  &\textcolor{blue}{-1.08} \\
            Track queries       &53.09  &\textcolor{blue}{-4.57}  &50.45  &\textcolor{blue}{-3.88}  \\
            \bottomrule
        \end{tabular}
        \vspace{-0.9em}
        \label{tab:ablation_study}
    \end{center}
    \vspace{-2mm}
\end{table}

\section{Conclusion}
\label{sec:conclusion}

In this work, we tackle the critical yet challenging task of 3D LN detection from a new tracking perspective and propose a novel LN tracking transformer, LN-Tracker, for joint end-to-end detection and 3D instance association. 
Built upon DETR-based detector, LN-Tracker decouples transformer decoder's query into the track and detection groups, where the track query autoregressively follows previously tracked LN instances along the z-axis of a CT scan. We design a new transformer decoder with masked attention module to align track query's content to the context of current slice, meanwhile preserving detection query’s high accuracy in current slice. An inter-slice similarity loss is introduced to encourage the cohesive LN association between slices. Extensive evaluation demonstrates LN-Tracker's superior performance, with at least $2.7\%$ gain in average sensitivity when compared to 3D and 2.5D detectors. Further validation on public lung nodule (LIDC) and prostate tumor (ProstateX) detection confirms the generalizability of LN-Tracker as it achieves top performance on both tasks.
{
    \small
    \bibliographystyle{ieeenat_fullname}
    \bibliography{main}
}

\setcounter{figure}{0}
\clearpage
\setcounter{page}{1}
\maketitlesupplementary

\section{Implementation Details}
\label{sec:imp}
\subsection{2.5D backbone}
As noted in \cite{yan2019mulan,yu2025effective}, utilizing 3D contextual information is crucial for accurately differentiating lymph nodes (LNs) from other tube-shaped structures like vessels and the esophagus. However, the direct implementation of 3D CNNs can be memory-intensive and often lacks pre-trained weights. To address this challenge, we incorporate the 2.5D feature fusion layer \cite{yu2025effective} into ResNet50, serving as the backbone for our LN-Tracker and other comparing methods. This approach allows us to leverage the rich 3D context information across slices.
\subsection{Transformer encoder and decoder}
The overall architecture of LN-Tracker's Transformer encoder and decoder is based on a unified DETR-like object detection and segmentation framework, Mask DINO \cite{li2023mask}. As in Mask DINO, we use a 6-layer Transformer encoder, a 9-layer Transformer decoder, and multiple shared FFN prediction heads across different decoder layers. In addition, we also introduce query denoising branch \cite{li2022dn,zhang2022dino,li2023mask,yu2025effective} to accelerate convergence and improve performance. Track queries are fed autoregressively from the previous slice output embeddings of the last Transformer decoder layer (before the final FFN prediction head). When decoding track and detection queries in each decoder layer, we apply an attention mask for the self-attention operation (\ie, masked attention), to prevent the information leakage from track queries to detection queries.

\subsection{Loss function}
\begin{figure}[t]
   \begin{center}
      \includegraphics[width=0.9\linewidth]{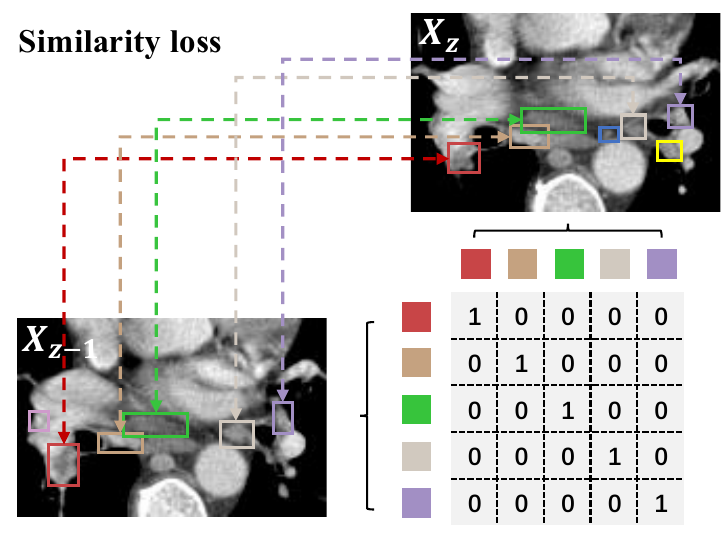}
   \end{center}
   \caption{Illustration of the similarity loss between track queries from adjacent slices. 1 represents the same LN and 0 represents different LN. }
   \label{fig:sim_loss}
\end{figure}

We use the focal loss with for classification and L1 loss and GIoU loss for box regression. The loss weights  $\lambda_{cls}, \lambda_{box}$ are set to 1.0 and 2.0, respectively. We also propose the similarity loss, as shown in Figure \ref{fig:sim_loss}, to encourage track queries belonging to the same instance across slices to be close in the latent space. As in DETR \cite{carion2020end}, we add auxiliary losses except similarity loss after each decoder layer. 

\subsection{Data pre-processing and augmentation}
For all LN datasets, we normalize the 3D CT volumes to a resolution of 0.8×0.8×2 mm and clip the intensity values to the range of [-200, 300]. In the case of the LIDC-IDRI dataset, we similarly normalize the 3D CT volumes to a resolution of 0.8×0.8×2 mm, but clip the intensity values to the range of [-1500, 500]. For the ProstateX dataset, we normalize the 3D T2 MRI volumes to a resolution of 0.5×0.5×3 mm and clip the intensity values to the range of [0, 800]. During training, we randomly sample pairs of adjacent slices from the 3D volumes and apply consistent augmentations to both slices in each pair. These augmentations include random scaling, cropping, rotation, intensity scaling, and gamma adjustment.

\begin{figure*}[t]
   \begin{center}
      \includegraphics[width=1.0\linewidth]{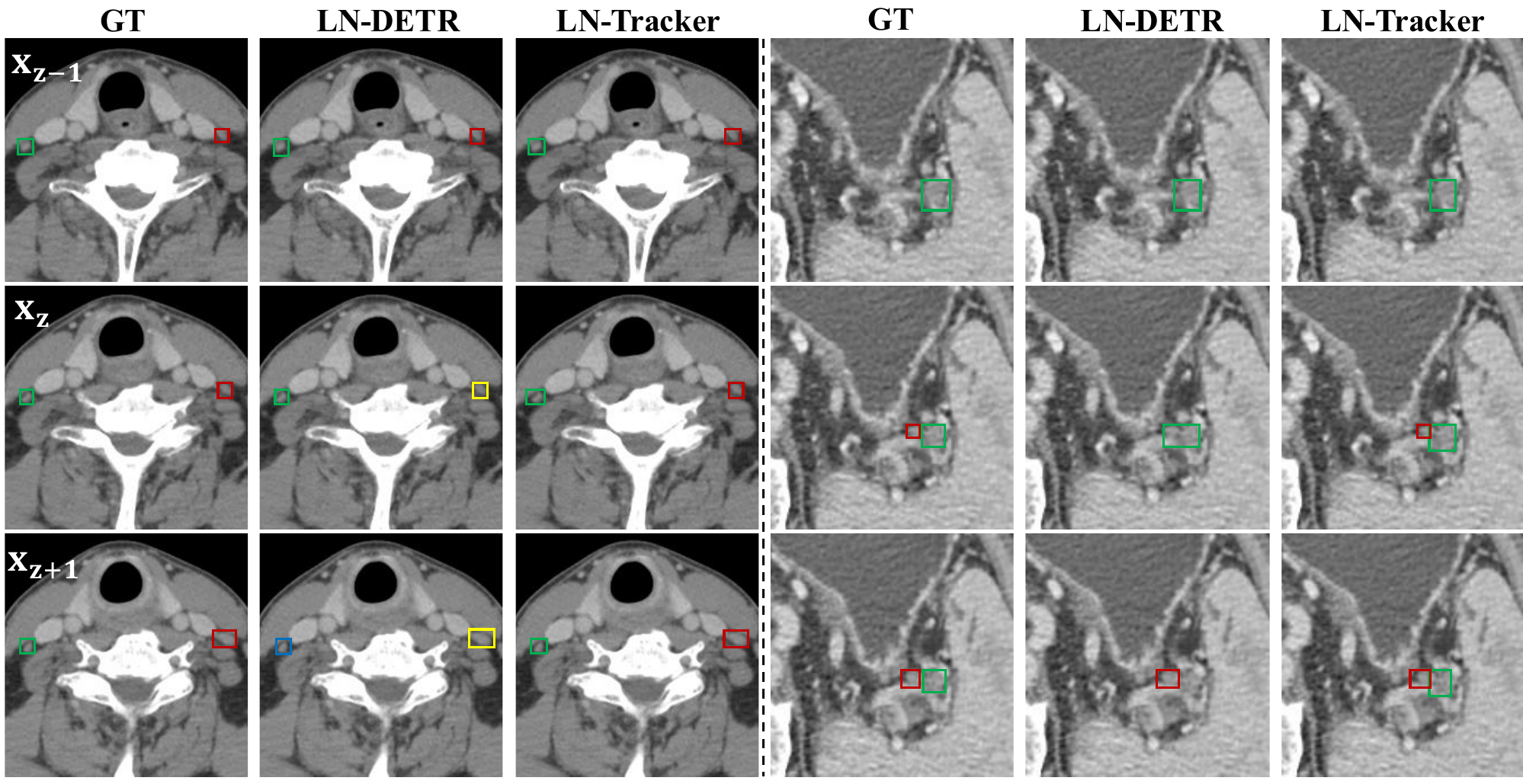}
   \end{center}
   \caption{Some qualitative results across consecutive slices $x_{z-1},x_z,x_{z+1}$. In the GT, boxes of the same color represent the same LN instance across slices. Missing boxes or color changes indicate missed detections or inconsistent associations between slices. }
   \label{fig:sup_qualitative}
\end{figure*}

\subsection{Details of comparing methods}
\noindent
\textbf{nnUNet.} nnUNet~\cite{isensee2021nnu} is an open-source deep learning framework designed specifically for 3D medical image segmentation, offering an automated and adaptable solution for a variety of image modalities and datasets. In our experiments, we first train nnUNet to segment the foreground region and then further decompose this region into multiple components, following \cite{guo2022thoracic}. For each component, we can easily determine its 3D bounding box by identifying the minimum and maximum coordinates along each dimension (x, y, z).

\noindent
\textbf{nnDetection.} nnDetection~\cite{baumgartner2021nndetection} is a self-configuring approach tailored for 3D medical object detection. It achieves results that are on par with or even superior to state-of-the-art methods across various public medical detection tasks. Upon completion of the training, we can directly obtain the 3D bounding box outputs along with their corresponding confidence scores.

\noindent
\textbf{MULAN.} MULAN~\cite{yan2019mulan} is a multi-task universal lesion analysis network for joint detection, tagging, and segmentation of lesions in a variety of body parts, which greatly extends existing work of single-task lesion analysis on specific body parts. MULAN is based on an improved Mask R-CNN framework with three head branches and a 3D feature fusion strategy. Here, we first train MULAN for slice-level LN detection, and then merging the 2D results to 3D boxes as \cite{yan2020learning}.

\noindent
\textbf{LN-DETR.} LN-DETR~\cite{yu2025effective} is a LN detection transformer developed based on Mask DINO, incorporating location-debiased query selection and contrastive query learning to enhance the representation capability of LN queries. This model has demonstrated impressive performance on both the LN detection task and the DeepLesion benchmark~\cite{yan2018deeplesion}. Here, we first train LN-DETR for slice-level LN detection, and subsequently merge the 2D results into 3D bounding boxes as \cite{yan2020learning}.

\noindent
\textbf{Mask2Former.} Mask2Former~\cite{cheng2022masked} utilizes a DETR-like architecture and is capable of addressing various image segmentation tasks, including panoptic, instance, and semantic segmentation. In this approach, we leverage its capability for slice-level instance segmentation to obtain 2D bounding boxes from the generated instance masks. These 2D boxes are then merged to form 3D bounding boxes, as described above.

\noindent
\textbf{VITA.} VITA~\cite{heo2022vita} is a straightforward video instance segmentation framework built upon Mask2Former. It begins by utilizing Mask2Former to aggregate frame-level object contexts into object tokens. Subsequently, VITA achieves video-level understanding by associating these frame-level object tokens, without relying on spatio-temporal backbone features. The resulting video-level instance tokens can be directly employed for video instance segmentation. Thus, we treat the 3D image volume as a video, with the z-axis representing the temporal axis. Consequently, we can directly obtain the 3D instance segmentation mask from the output.

\noindent
\textbf{GenVIS.} GenVIS~\cite{heo2023generalized} is a novel generalized video instance segmentation framework that sequentially associate clip-wise predictions.
Beyond the score-based associations, it take a further step to design a non-heuristic clip-wise association under the motivation of eliminating the barriers between training and inference. Like the above VITA, we can directly obtain the 3D instance segmentation mask from the output.

\noindent
\textbf{LN-DETR+SAM2.} SAM2~\cite{ravi2024sam2segmentimages} is a foundation model towards solving promptable visual segmentation in images and videos. With points, boxes or mask as prompts, SAM2 can accurately segment the object foreground within an image and track the object across frames in a video. Leveraging its object tracking capability, we employ slice-level LN predictions from the fully trained LN-DETR as prompts for SAM2 to get the whole 3D LN results like Grounded-SAM2\footnote{\url{https://github.com/IDEA-Research/Grounded-SAM-2}}.


\section{More visualization}
\label{sec:vis}
Figure \ref{fig:sup_qualitative} presents qualitative comparisons with the leading LN detection method, LN-DETR. In the first three columns, while LN-DETR successfully detects all LNs in consecutive slices, its heuristic IoU-based stacking fails to associate the same instance across slices. In contrast, our LN-Tracker consistently aggregates predictions across slices, producing cohesive 3D results. In the last three columns, LN-DETR incorrectly merges two clustered LNs into one in the second row, leading to a failed association in the following slice. LN-Tracker, however, maintains accurate and consistent detection and tracking throughout the sequence.

{
}
\end{document}